\documentclass[11pt,a4paper]{article}

\usepackage[hyperref]{acl2020}
\usepackage{booktabs}
\usepackage{times}
\usepackage{graphicx}
\usepackage{tikz}
\usepackage{latexsym}
\usepackage{pgfplots}
\usepackage{booktabs}
\usepackage{amsmath}
\usepackage{url}
\usepackage[encapsulated]{CJK}
\usepackage{microtype}

\newcommand{\ignore}[1]{}

\aclfinalcopy

\title{Rapid Adaptation of BERT for Information Extraction\\ on Domain-Specific Business Documents}

\author{{\bf Ruixue Zhang},$^{1,2}$ {\bf Wei Yang},$^{1}$ {\bf Luyun Lin},$^{1}$ {\bf Zhengkai Tu},$^{1}$ {\bf Yuqing Xie},$^{1,2}$ \\ 
{\bf Zihang Fu},$^{1}$ {\bf Yuhao Xie},$^{1}$ {\bf Luchen Tan},$^{1}$ {\bf Kun Xiong},$^{1}$ and {\bf Jimmy Lin}$^{1,2}$\vspace{0.1cm}\\
    $^1$ RSVP.ai\quad
    $^2$ David R. Cheriton School of Computer Science, University of Waterloo\quad
}

\date{}

\begin{document}

\maketitle

\begin{abstract}
Techniques for automatically extracting important content elements from business documents such as contracts, statements, and filings have the potential to make business operations more efficient.
This problem can be formulated as a sequence labeling task, and we demonstrate the adaption of BERT to two types of business documents:\ regulatory filings and property lease agreements.
There are aspects of this problem that make it easier than ``standard'' information extraction tasks and other aspects that make it more difficult, but on balance we find that modest amounts of annotated data (less than 100 documents) are sufficient to achieve reasonable accuracy.
We integrate our models into an end-to-end cloud platform that provides both an easy-to-use annotation interface as well as an inference interface that allows users to upload documents and inspect model outputs.
\end{abstract}

\section{Introduction}

{\it Business documents} broadly characterize a large class of documents that are central to the operation of business.
These include legal contracts, purchase orders, financial statements, regulatory filings, and more.
Such documents have a number of characteristics that set them apart from the types of texts that most NLP techniques today are designed to process (Wikipedia articles, news stories, web pages, etc.):\ They are heterogeneous and frequently contain a mix of both free text as well as semi-structured elements (tables, headings, etc.).
They are, by definition, domain specific, often with vocabulary, phrases, and linguistic structures (e.g., legal boilerplate and terms of art) that are rarely seen in general natural language corpora.

Despite these challenges, there is great potential in the application of NLP technologies to business documents.
Take, for example, contracts that codify legal agreements between two or more parties.
Organizations (particularly large enterprises) need to monitor contracts for a range of tasks, a process that can be partially automated if certain content elements can be extracted from the contracts themselves by systems~\cite{Chalkidis:2017:ECE:3086512.3086515}.
In general, if we are able to extract structured entities from business documents, these outputs can be better queried and manipulated, potentially facilitating more efficient business operations.

In this paper, we present BERT-based models for extracting content elements from two very different types of business documents:\ regulatory filings and property lease agreements.
Given the success of deep transformer-based models such as BERT~\cite{devlin-etal-2019-bert} and their ability to handle sequence labeling tasks, adopting such an approach seemed like an obvious starting point.
In this context, we are primarily interested in two questions:
First, how data efficient is BERT for fine-tuning to new specialized domains?
Specifically, how much annotated data do we need to achieve some (reasonable) level of accuracy?
This is an important question due to the heterogeneity of business documents; it would be onerous if organizations were required to engage in large annotation efforts for every type of document.
Second, how would a BERT model pre-trained on general natural language corpora perform in specific, and potentially highly-specialized, domains?

There are aspects of this task that make it both easier and more difficult than ``traditional'' IE.
Even though they are expressed in natural language, business documents frequently take constrained forms, sometimes even ``template-like'' to a certain degree.
As such, it may be easy to learn cue phrases and other fixed expressions that indicate the presence of some element (i.e., pattern matching).
On the other hand, the structure and vocabulary of the texts may be very different from the types of corpora modern deep models are trained on; for example, researchers have shown that models for processing the scientific literature benefit immensely from pre-training on scientific articles~\cite{beltagy2019scibert,Nogueira_etal_arXiv2020a}.
Unfortunately, we are not aware of any large, open corpora of business documents for running comparable experiments.

The contribution of our work is twofold:\
From the scientific perspective, we begin to provide some answers to the above questions.
With two case studies, we find that a modest amount of domain-specific annotated data (less than 100 documents) is sufficient to fine-tune BERT to achieve reasonable accuracy in extracting a set of content elements.
From a practical perspective, we showcase our efforts in an end-to-end cloud platform that provides an easy-to-use annotation interface as well as an inference interface that allows users to upload documents and inspect the results of our models.

\section{Approach}

Within the broad space of business documents, we have decided to focus on two specific types:\ regulatory filings and property lease agreements.
While our approach is not language specific, all our work is conducted on Chinese documents.
In this section, we first describe these documents and our corpora, our sequence labeling model, and finally our evaluation approach.

\subsection{Datasets}

\noindent {\bf Regulatory Filings.}
We focused on a specific type of filing:\ disclosures of pledges by shareholders when their shares are offered up for collateral. 
These are publicly accessible and were gathered from the database of a stock exchange in China.
We observe that most of these announcements are fairly formulaic, likely generated by templates.
However, we treated them all as natural language text and did not exploit this observation; for example, we made no explicit attempt to induce template structure or apply clustering---although such techniques would likely improve extraction accuracy.
In total, we collected and manually annotated 150 filings, which were divided into training, validation, and test sets with a 6:2:2 split.
Our test corpus comprises 30 regulatory filings.
Table~\ref{tab:finance} enumerates the seven content elements that we extract.

\smallskip \noindent {\bf Property Lease Agreements.}
These contracts mostly follow a fixed ``schema'' with a certain number of prescribed elements (leaseholder, tenant, rent, deposit, etc.); Table~\ref{tab:contract} enumerates the eight elements that our model extracts.
Since most property lease agreements are confidential, no public corpus for research exists, and thus we had to build our own.
To this end, we searched the web for publicly-available templates of property lease agreements and found 115 templates in total.
For each template, we manually generated one, two, or three instances, using a fake data generator tool\footnote{\url{https://github.com/joke2k/faker}} to fill in the missing content elements such as addresses.
In total, we created (and annotated) 223 contracts by hand.
This corpus was further split into training, validation, and test data with a 6:2:2 split.
Our test set contains 44 lease agreements, 11 of which use templates that are not seen in the training set.
We report evaluation over both the full test set and on only these unseen templates; the latter condition specifically probes our model's ability to generalize.

\subsection{Model}

An obvious approach to content element extraction is to formulate the problem as a sequence labeling task.
Prior to the advent of neural networks, Conditional Random Fields (CRFs) \cite{10.3115/1219840.1219885, 6707709} represented the most popular approach to this task. Starting from a few years ago, neural networks have become the dominant approach, starting with RNNs~\cite{DBLP:journals/corr/HuangXY15, kurata2016leveraging, liu2016attention,
hakkani-tr2016multi-domain, Wang2018ABB, gangadharaiah-narayanaswamy-2019-joint}.
Most recently, deep transformer-based models such as BERT represent the state of the art in this task~\cite{devlin-etal-2019-bert,DBLP:journals/corr/abs-1902-10909, zhang-etal-2019-joint} .
We adopt the sequence labeling approach of~\citet{devlin-etal-2019-bert}, based on annotations of our corpus using a standard BIO tagging scheme with respect to the content elements we are interested in.

We extend BERT Base-Chinese (12-layer, 768-hidden, 12-heads, 110M parameters) for sequence labeling.
All documents are segmented into paragraphs and processed at the paragraph level (both training and inference); this is acceptable because we observe that most paragraphs are less than 200 characters.
The input sequences are segmented by the BERT tokenizer, with the special [CLS] token inserted at the beginning and the special [SEP] token added at the end.
All inputs are then padded to a length of 256 tokens.
After feeding through BERT, we obtain the hidden state of the final layer, denoted as ($h_{1}$, $h_{2}$, ... $h_{N}$) where $N$ is the max length setting.
We add a fully-connected layer and softmax on top, and the final prediction is formulated as:
\begin{equation}
y_{n}  = \textrm{softmax}(Wh_{n}  + b),  n \in {1, 2, ... N}
\end{equation}
\noindent where ${W}$ represents the parameter of the fully-connected layer and ${b}$ is the bias. The learning objective is to maximize 
\begin{equation}
P(y | x) =  \prod_{i=1}^{N}P(y_{n} | x),  n \in {1, 2, ... N}
\end{equation}
For simplicity, we assume that all tokens can be predicted independently.
For model training, we set the max sequence length to 256, the learning rate to ${10^{-4}}$, and run the model for 8 epochs.
We use all other default settings in the TensorFlow implementation of BERT.

\begin{CJK}{UTF8}{gbsn}
\begin{table*}[t]
	\centering
	\begin{small}
	\begin{tabular}{l|ccc}
		\toprule
		& F$_1$ & Precision & Recall \\
		\midrule
		All                  & 0.83     & 0.75     & 0.92   \\
		\midrule
		本次质押占公司总股本比例详情 \\ {\small \% of pledged shares in total shares issued by the corporation}       & 0.72     & 0.58     & 0.95   \\
		股东名称详情 \\ {\small Name of the shareholder(s)}               & 0.73     & 0.69     & 0.77   \\
		股东简称详情 \\ {\small Abbreviation or initials of the shareholder(s)}              & 0.85     & 0.77     & 0.94   \\
		该股东持股数量详情 \\ {\small Number of shares owned}           & 0.92     & 0.86     & 1.00      \\
		该股东持股比例详情      \\ {\small \% of shares owned}      & 0.88     & 0.79     & 0.99   \\
		该股东本次质押股份数量占其所持股份比例详情 \\ {\small \% of pledged shares in the shareholder's total share holdings} & 0.40      & 0.50      & 0.33   \\
		该股东本次质押股份数量详情 \\ {\small Number of the shareholder's pledged shares}        & 0.88     & 0.81     & 0.97   \\
		\bottomrule
	\end{tabular}
	\end{small}
	\caption{Evaluation results on the test set of our regulatory filings corpus.} 
	\label{tab:finance}
\end{table*}
\end{CJK}

\begin{CJK}{UTF8}{gbsn}
\begin{table*}[t]
	\centering
	\begin{small}
	\begin{tabular}{l|ccc|ccc}
		\toprule
		& \multicolumn{3}{c|}{\bf All}      & \multicolumn{3}{c}{\bf Unseen Templates} \\
		& F$_1$ & Precision & Recall & F$_1$     & Precision    & Recall   \\
		\midrule
		All             & 0.83     & 0.82      & 0.80   & 0.73        & 0.74         & 0.72     \\
		\midrule
		甲方 (Leaseholder)          & 0.91     & 0.91      & 0.86   & 0.82        & 0.83         & 0.82     \\
		乙方 (Tenant)
		& 0.94     & 0.94      & 0.94   & 0.83        & 0.90         & 0.77     \\
		租金 (Rent)            & 0.76     & 0.75      & 0.77   & 0.59        & 0.59         & 0.59     \\
		押金 (Deposit)         & 0.62     & 0.60      & 0.59   & 0.42        & 0.42         & 0.43     \\
		合同年限 (Lease Term) & 0.75     & 0.70      & 0.72   & 0.57        & 0.52         & 0.62     \\
		合同开始时间 (Start Date)      & 0.80     & 0.79      & 0.79   & 0.82        & 0.82         & 0.82     \\
		合同结束时间 (End Date)        & 0.81     & 0.79      & 0.79   & 0.82        & 0.82         & 0.82     \\
		合同签订时间 (Date of Signature)  & 0.91     & 0.91      & 0.89   & 0.84        & 0.89         & 0.80     \\
		\bottomrule
	\end{tabular}
	\end{small}
	\caption{Evaluation results on the test set of our property lease agreements corpus.}
	\label{tab:contract}
\end{table*}
\end{CJK}

\begin{CJK}{UTF8}{gbsn}
\begin{table*}[t]
\centering
\begin{small}
\begin{tabular}{l|l}
\toprule
\noalign{\vskip 2mm}
\multicolumn{2}{p{15cm}}{截至本公告日，上海览海共持有公司 304,642,913 股股票，占公司总股本的 35.05\%。} \\
\multicolumn{2}{p{15cm}}{As of the date of this announcement, Shanghai Lanhai holds a total of 304,642,913 shares of the company, accounting for  35.05\% of the company's total shares outstanding.} \\
\noalign{\vskip 2mm}
\toprule
{\bf Content Element} & {\bf Value} \\
\midrule
股东简称详情  & 上海览海 \\
Abbreviation or initials of the shareholder(s) & Shanghai Lanhai \\
该股东持股数量详情 & 304,642,913 \\
Number of shares owned & \\
该股东持股比例详情 & 35.05\% \\
\% of shares owned & \\
\toprule
\noalign{\vskip 2mm}
\multicolumn{2}{p{15cm}}{第二条租赁期限 五、租赁期2年，自2019年1月1日起至2020年12月31日止。第三条租金及其它费用 六、合同有效年度租金共计为160000元（人民币） } \\
\multicolumn{2}{p{15cm}}{2.~Term of lease 5)~The lease term is 2 years, from January 1,  2019 to December 31, 2020.  3.~Rent and other expenses 6)~The effective annual rent of the contract is 160,000 Yuan(RMB)} \\
\noalign{\vskip 2mm}
\toprule
{\bf Content Element} & {\bf Value} \\
\midrule
合同年限 & 租赁期2年 \\ 
Lease Term & 2 year lease \\ 
合同开始时间 & 2019年1月1日 \\ 
Start Date & January 1, 2019 \\ 
合同结束时间 & 2020年12月31日 \\ 
End Date & December 31, 2020 \\ 
租金 & 年度租金共计160000元（人民币） \\
Rent & 160,000 Yuan(RMB)\\             
\bottomrule
\end{tabular}
\end{small}
\caption{Excerpts from a regulatory filing (top) and a property lease agreement (bottom) illustrating a few of the content elements that our models extract.}
\label{tab:label_exampleas}
\end{table*}
\end{CJK}

\subsection{Inference and Evaluation}

At inference time, documents from the test set are segmented into paragraphs and fed into the fine-tuned BERT model one at a time.
Typically, sequence labeling tasks are evaluated in terms of precision, recall, and F$_1$ at the entity level, per sentence.
However, such an evaluation is inappropriate for our task because the content elements represent properties of the entire document as a whole, not individual sentences.

Instead, we adopted the following evaluation procedure:\
For each content element type (e.g., ``tenant''), we extract {\it all} tagged spans from the document, and after deduplication, treat the entities as a set that we then measure against the ground truth in terms of precision, recall, and F$_1$.
We do this because there may be multiple ground truth entities and BERT may mark multiple spans in a document with a particular entity type.
Note that the metrics are based on {\it exact} matches---this means that, for example, if the extracted entity has an extraneous token compared to a ground truth entity, the system receives no credit.

\section{Results}

Our main results are presented in Table~\ref{tab:finance} on the test set of the regulatory filings and in Table~\ref{tab:contract} on the test set of the property lease agreements; F$_1$, precision, and recall are computed in the manner described above.
We show metrics across all content elements (micro-averaged) as well as broken down by types.
For the property lease agreements, we show results on all documents (left) and only over those with unseen templates (right).
Examining these results, we see that although there is some degradation in effectiveness between all documents and only unseen templates, it appears that BERT is able to generalize to previously-unseen expressions of the content elements.
Specifically, it is {\it not} the case that the model is simply memorizing fixed patterns or key phrases---otherwise, we could just craft a bunch of regular expression patterns for this task.
This is a nice result that shows off the power of modern neural NLP models.

Overall, we would characterize our models as achieving reasonable accuracy, comparable to extraction tasks in more ``traditional'' domains, with modest amounts of training data.
It does appear that with fine tuning, BERT is able to adapt to the linguistic characteristics of these specialized types of documents.
For example, the regulatory filings have quite specialized vocabulary and the property lease agreements have numeric heading structures---BERT does not seem to be confused by these elements, which for the most part do not appear in the texts that the model was pre-trained on.
Naturally, accuracy varies across different content elements:
For the rental agreements, entities such as leaseholder, tenant, start date, and end date perform much better than others.
For the regulatory filing, the model performs well on all content elements except for one; there were very few examples of ``\% of pledged shares in the shareholder's total share holdings'' in our training data, and thus accuracy is very low despite the fact that percentages are straightforward to identify.
It seems that ``easy'' entities often have more fixed forms and are quite close to entities that the model may have encountered during pre-training (e.g., names and dates).
In contrast, ``difficult'' elements are often domain-specific and widely vary in their forms.

\begin{figure}[t]
	\centering
	\includegraphics[width=1.0\linewidth]{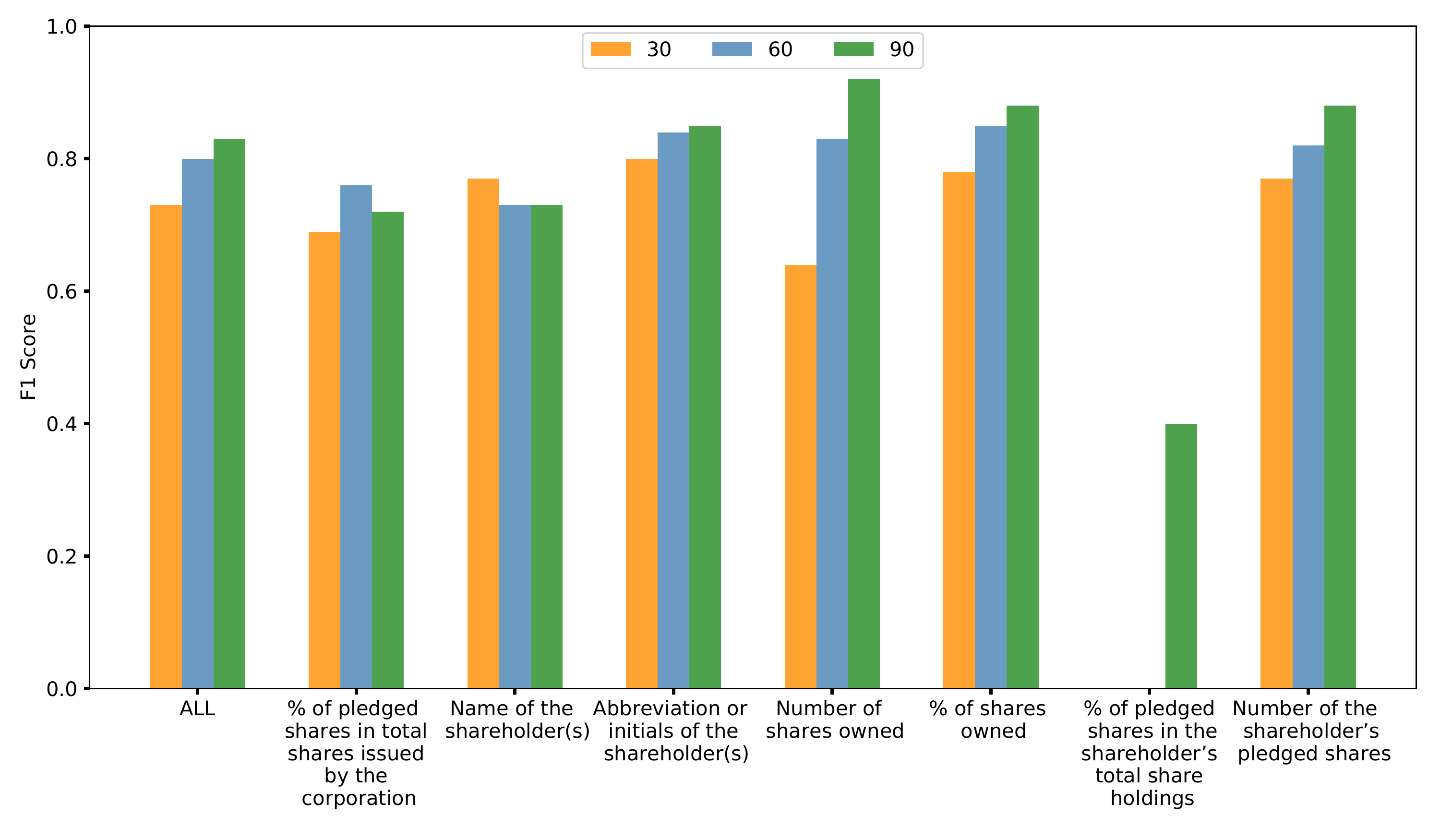}
	\caption{Effects of training data size on F$_1$ for regulatory filings.}
	\label{fig:finance_f1_score}
\end{figure}

\begin{figure}[t]
	\centering
	\includegraphics[width=1.0\linewidth]{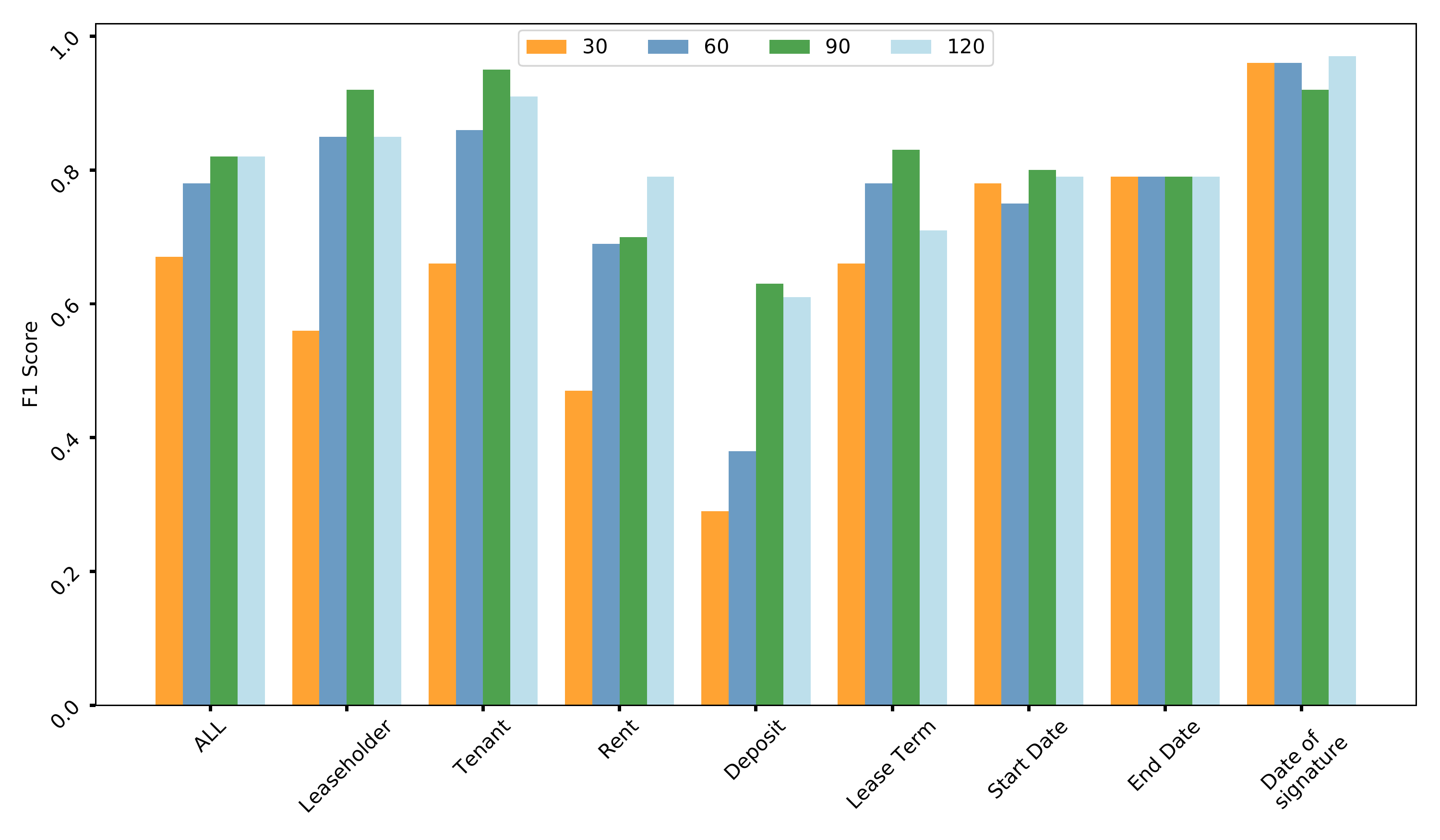}
	\caption{Effects of training data size on F$_1$ for property lease agreements.}
	\label{fig:contract_f1_score}
\end{figure}

How data efficient is BERT when fine tuning on annotated data?
We can answer this question by varying the amount of training data used to fine tune the BERT models, holding everything else constant.
These results are shown in Figure~\ref{fig:finance_f1_score} for the regulatory filings (30, 60, 90 randomly-selected documents) and in Figure~\ref{fig:contract_f1_score} for the property lease agreements (30, 60, 90, 120 randomly-selected documents); in all cases, the development set is fixed.
For brevity, we only show F$_1$ scores, but we observe similar trends for the other metrics.
For both document types, it seems like 60--90 documents are sufficient to achieve F$_1$ on par with using all available training data.
Beyond this point, we hit rapidly diminishing returns.
For a number of ``easy'' content elements (e.g., dates in the property lease agreements), it seems like 30 documents are sufficient to achieve good accuracy, and more does not appear to yield substantial improvements.
Note that in a few cases, training on {\it more} data actually decreases F$_1$ slightly, but this can be attributed to noise in the sampling process.

\begin{figure*}[t]
\centering
\includegraphics[width=0.8\linewidth]{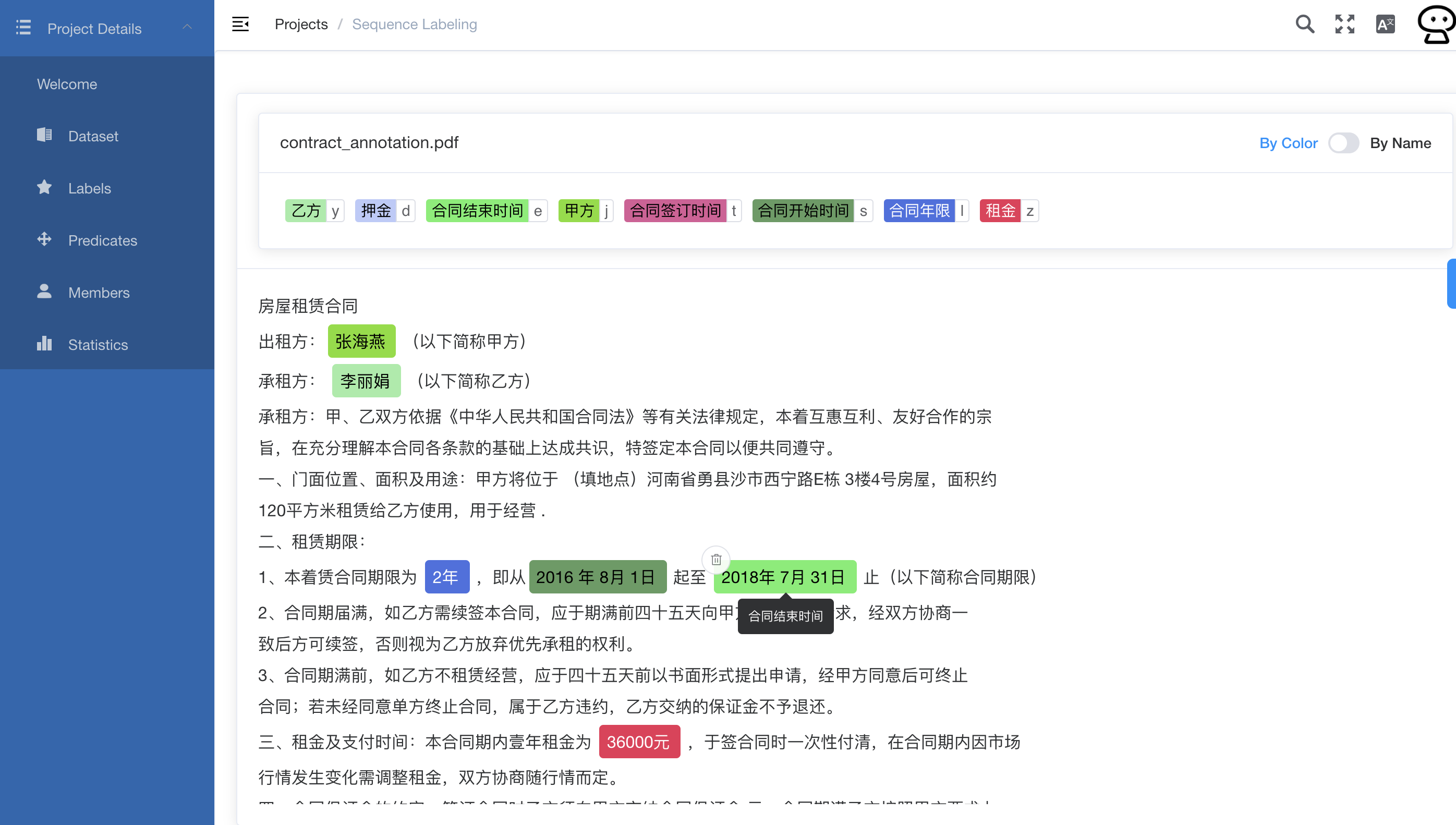}
\caption{Screenshot of our annotation interface.}
\label{fig:interface:annotation}
\end{figure*}

\begin{figure*}[t]
\centering
\includegraphics[width=0.8\linewidth]{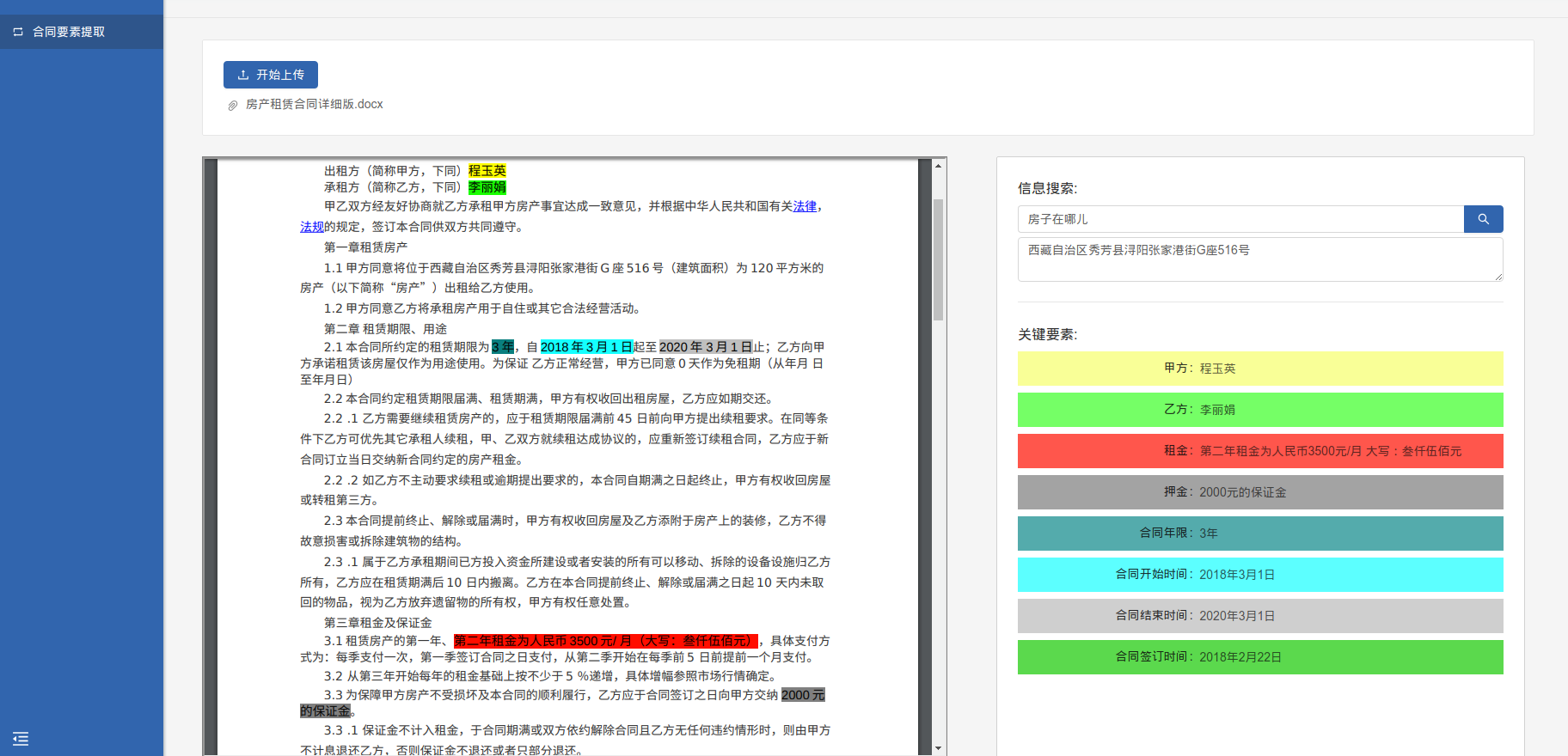}
\caption{Screenshot of our inference interface.}
\label{fig:interface:inference}
\end{figure*}

Finally, in Table~\ref{tab:label_exampleas} we show an excerpt from each type of document along with the content elements that are extracted by our BERT models.
We provide both the original source Chinese texts as well as English translations to provide the reader with a general sense of the source documents and how well our models behave.

\section{Cloud Platform}

All the capabilities described in this paper come together in an end-to-end cloud-based platform that we have built.
The platform has two main features:\ 
First, it provides an annotation interface that allows users to define content elements, upload documents, and annotate documents; a screenshot is shown in Figure~\ref{fig:interface:annotation}.
We have invested substantial effort in making the interface as easy to use as possible; for example, annotating content elements is as easy as selecting text from the document.
Our platform is able to ingest documents in a variety of formats, including PDFs and Microsoft Word, and converts these formats into plain text before presenting them to the annotators.

The second feature of the platform is the ability for users to upload new documents and apply inference on them using a fine-tuned BERT model; a screenshot of this feature is shown in Figure~\ref{fig:interface:inference}.
The relevant content elements are highlighted in the document.

On the cloud platform, the inference module also applies a few simple rule-based modifications to post-process BERT extraction results.
For any of the extracted dates, we further applied a date parser based on rules and regular expressions to normalize and canonicalize the extracted outputs.
In the regulatory filings, we tried to normalize numbers that were written in a mixture of Arabic numerals and Chinese units (e.g., ``\begin{CJK}{UTF8}{gbsn}亿\end{CJK}'', the unit for $10^8$) and discarded partial results if simple rule-based rewrites were not successful.
In the property lease agreements, the contract length, if not directly extracted by BERT, is computed from the extracted start and end dates.
Note that these post processing steps were not applied in the evaluation presented in the previous section, and so the figures reported in Tables~\ref{tab:finance} and~\ref{tab:contract} actually under-report the accuracy of our models in a real-world setting.

\section{Conclusions}

This work tackles the challenge of content extraction from two types of business documents, regulatory filings and property lease agreements.
The problem is straightforwardly formulated as a sequence labeling task, and we fine-tune BERT for this application.
We show that our simple models can achieve reasonable accuracy with only modest amounts of training data, illustrating the power and flexibility of modern NLP models.
Our cloud platform pulls these models together in an easy-to-use interface for addressing real-world business needs.

\bibliographystyle{acl_natbib}
%\bibliography{biz-docs}

\end{document}